\documentclass[letterpaper, 10 pt, conference]{ieeeconf}  

\IEEEoverridecommandlockouts                              

\usepackage{graphics} 
\usepackage{epsfig} 
\usepackage{times} 
\usepackage{amsmath} 
\usepackage{amsfonts} 
\usepackage{amssymb}  
\usepackage{stmaryrd}
\usepackage{mathtools}
\usepackage[dvipsnames]{xcolor}
\usepackage[caption=false,font=footnotesize]{subfig}
\usepackage{hyperref}

\newcommand{\pos}{\mathbf{q}}

\newcommand{\stateMCTS}{\textsl{s}}
\newcommand{\actionMCTS}{\textsl{a}}
\newcommand{\contact}{\mathsf{c}}
\newcommand{\patch}{\mathsf{p}}
\newcommand{\mode}{\mathsf{M}}
\newcommand{\Neeff}{n_{\text{ee}}}
\newcommand{\Npatch}{n_{\patch}}

\newcommand{\Nphase}{n_{\varphi}}
\newcommand{\mincnt}{n_{\min}}

\def\*#1{\mathbf{#1}}
\DeclarePairedDelimiter\norm{\lVert}{\rVert}

\title{\LARGE \bf
Simultaneous Contact Sequence and Patch Planning for Dynamic Locomotion
}

\author{Victor Dhédin$^{1}$, Haizhou Zhao$^{2}$, and Majid Khadiv$^{1}$
\thanks{This work was partially supported by the Huawei-TUM joint laboratory- Individual Project Agreements TC20231218038-2025-2 and TC20231218038-2025-3}
\thanks{$^{1}$Munich Institute of Robotics and Machine Intelligence (MIRMI), Technical University of Munich (TUM), Germany. {\tt\small firstname.lastname@tum.de}}
\thanks{$^{2}$Tandon School of Engineering, New York University
(NYU), USA. {\tt\small hz3862@nyu.edu}}
\thanks{$^{*}$\url{https://youtu.be/Is_NNQbW10c}}
}

\begin{document}

\maketitle
\thispagestyle{empty}
\pagestyle{empty}

\begin{abstract}
Legged robots have the potential to traverse highly constrained environments with agile maneuvers. However, planning such motions requires solving a highly challenging optimization problem with a mixture of continuous and discrete decision variables. In this paper, we present a full pipeline based on Monte-Carlo tree search (MCTS) and whole-body trajectory optimization (TO) to perform simultaneous contact sequence and patch selection on highly challenging environments. Through extensive simulation experiments, we show that our framework can quickly find a diverse set of dynamically consistent plans. We experimentally show that these plans are transferable to a real quadruped robot. We further show that the same framework can find highly complex acyclic humanoid maneuvers. To the best of our knowledge, this is the first demonstration of simultaneous contact sequence and patch selection for acyclic multi-contact locomotion using the whole-body dynamics of a quadruped (\href{https://youtu.be/Is_NNQbW10c}{video}$^{*}$).
\end{abstract}

\section{Introduction}

Contact interaction is at the heart of any locomotion planning and control problem. Therefore, methods to generate locomotion behaviors can be categorized based on how they handle contact. Contact-implicit methods solve the problem without explicitly planning for contact modes. The classic way of solving such a problem is through gradient-based trajectory optimization (TO) with complementarity constraints \cite{posa2014direct} that have recently become fast enough to enable closed-loop model predictive control (MPC) \cite{kurtz2023inverse,le2024fast,kim2025contact}. However, as it has long been argued and has been shown in \cite{nurkanovic2020limits}, mathematical programs with complementarity constraints (MPCC) suffer from getting stuck in poor local minima. 

Thanks to the recent advances in fast parallel simulations \cite{makoviychuk2021isaac,zakka2025mujoco}, there has been a recent effort towards sampling-based MPC \cite{howell2022predictive,alvarez2024real} to resolve the problem of local minima of MPCCs. However, this approach still has limited capability for high-dimensional and complex environments due to the large dimensionality of the sampling space \cite{kurtz2025generative}.
Another approach to solve the contact-implicit policy optimization problem is through the use of deep reinforcement learning (DRL) \cite{ha2024learning}. In contrast with MPC, DRL solves the policy optimization problem entirely offline. Furthermore, DRL requires heavy reward shaping and very long training time for locomotion in highly constrained environments \cite{zhang2024learning}.

\begin{figure}[!t]
    \vspace{2.5mm}
    \centering

    \subfloat{
        \reflectbox{
            \includegraphics[width=0.45\linewidth, trim=4cm 3cm 5cm 4.5cm, clip]{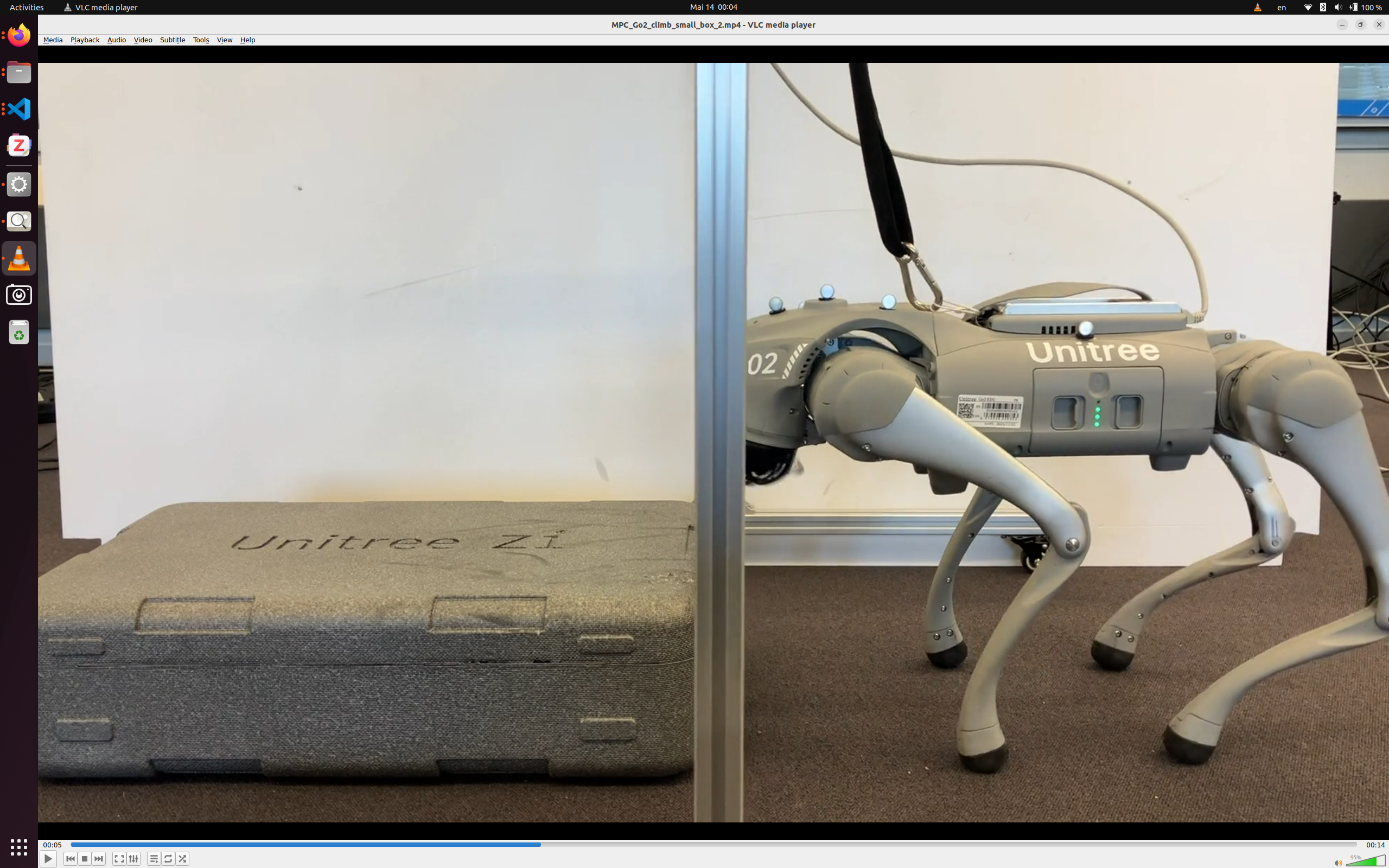}
        }
    }
    \hspace{-5mm}
    \subfloat{
        \reflectbox{
            \includegraphics[width=0.45\linewidth, trim=4cm 3cm 5cm 4.5cm, clip]{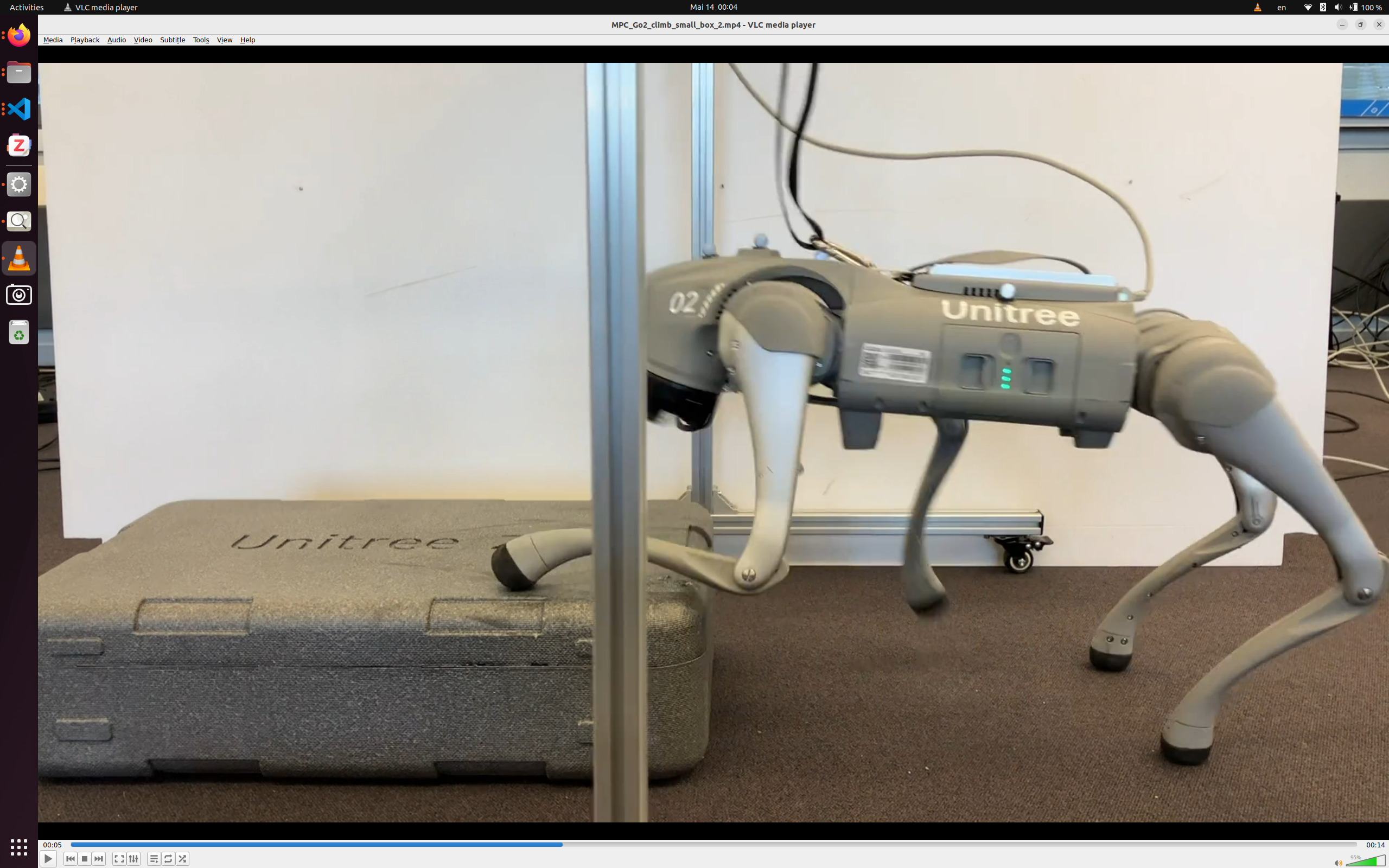}
        }
    }
    \\
    \vspace{-2.5mm}
    \subfloat{
        \reflectbox{
            \includegraphics[width=0.45\linewidth, trim=4cm 3cm 5cm 4.5cm, clip]{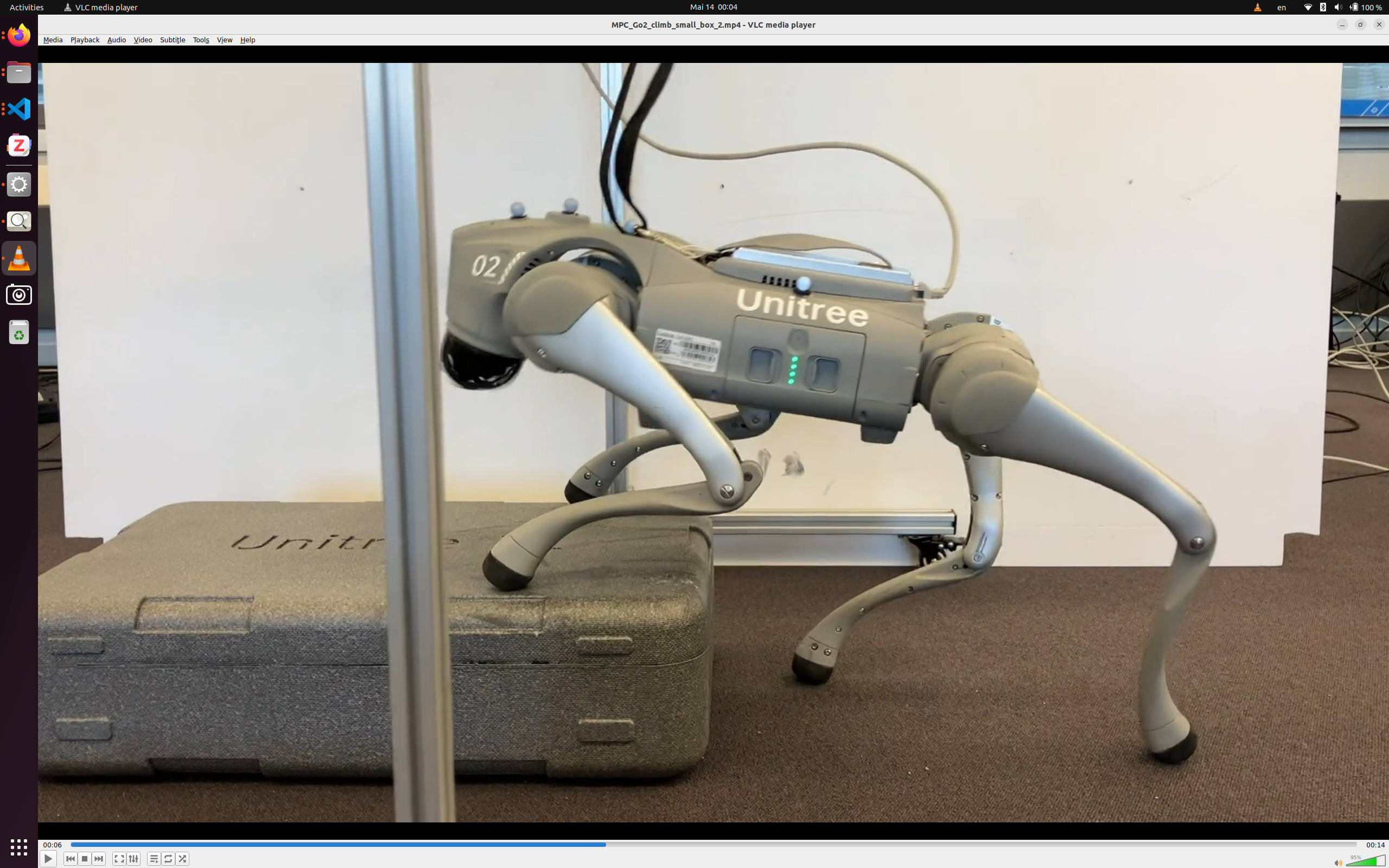}
        }
    }
    \hspace{-5mm}
    \subfloat{
        \reflectbox{
            \includegraphics[width=0.45\linewidth, trim=4cm 3cm 5cm 4.5cm, clip]{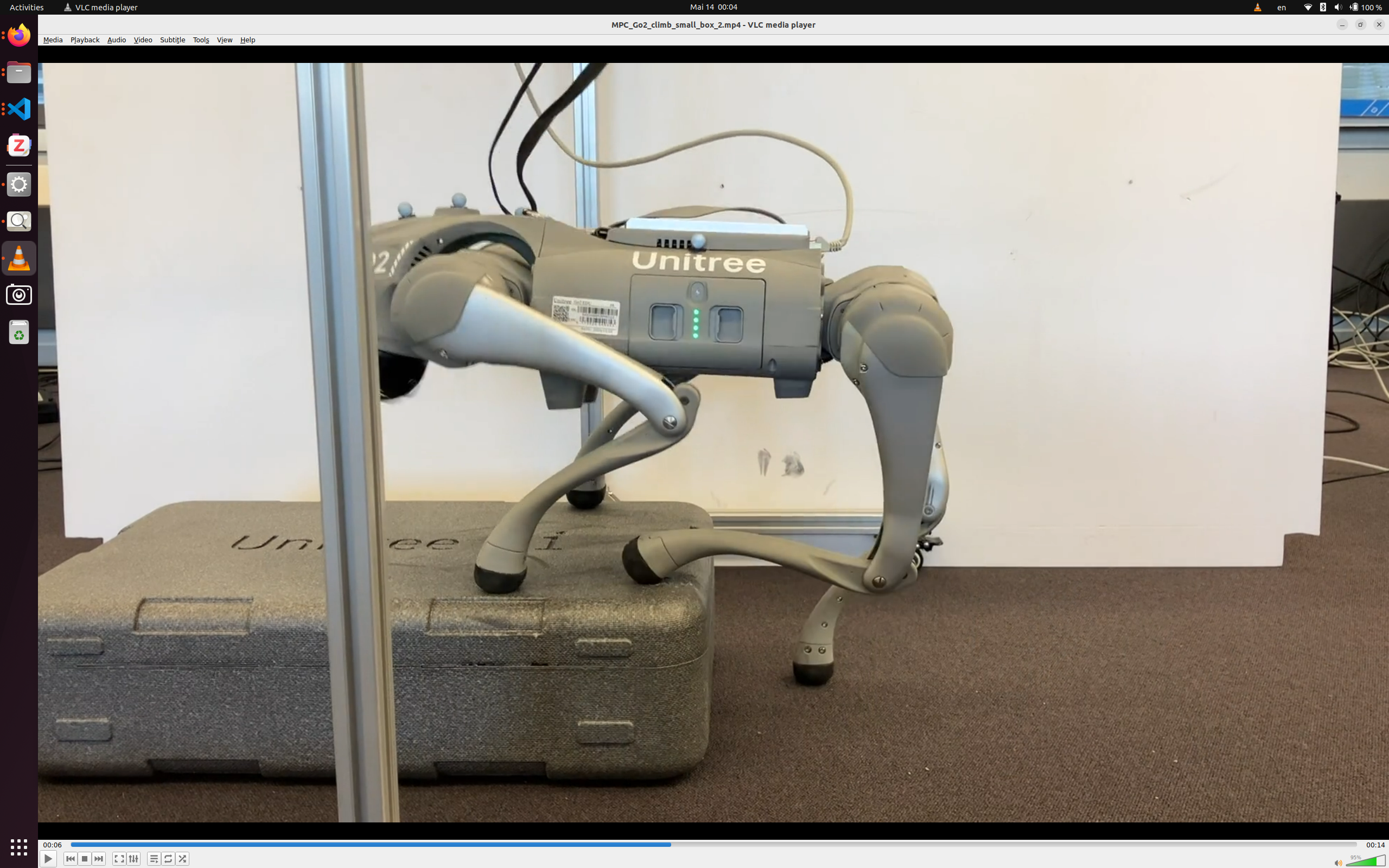}
        }
    }
    \\
    \vspace{-2.5mm}
    \subfloat{
        \reflectbox{
            \includegraphics[width=0.45\linewidth, trim=4cm 3cm 5cm 4.5cm, clip]{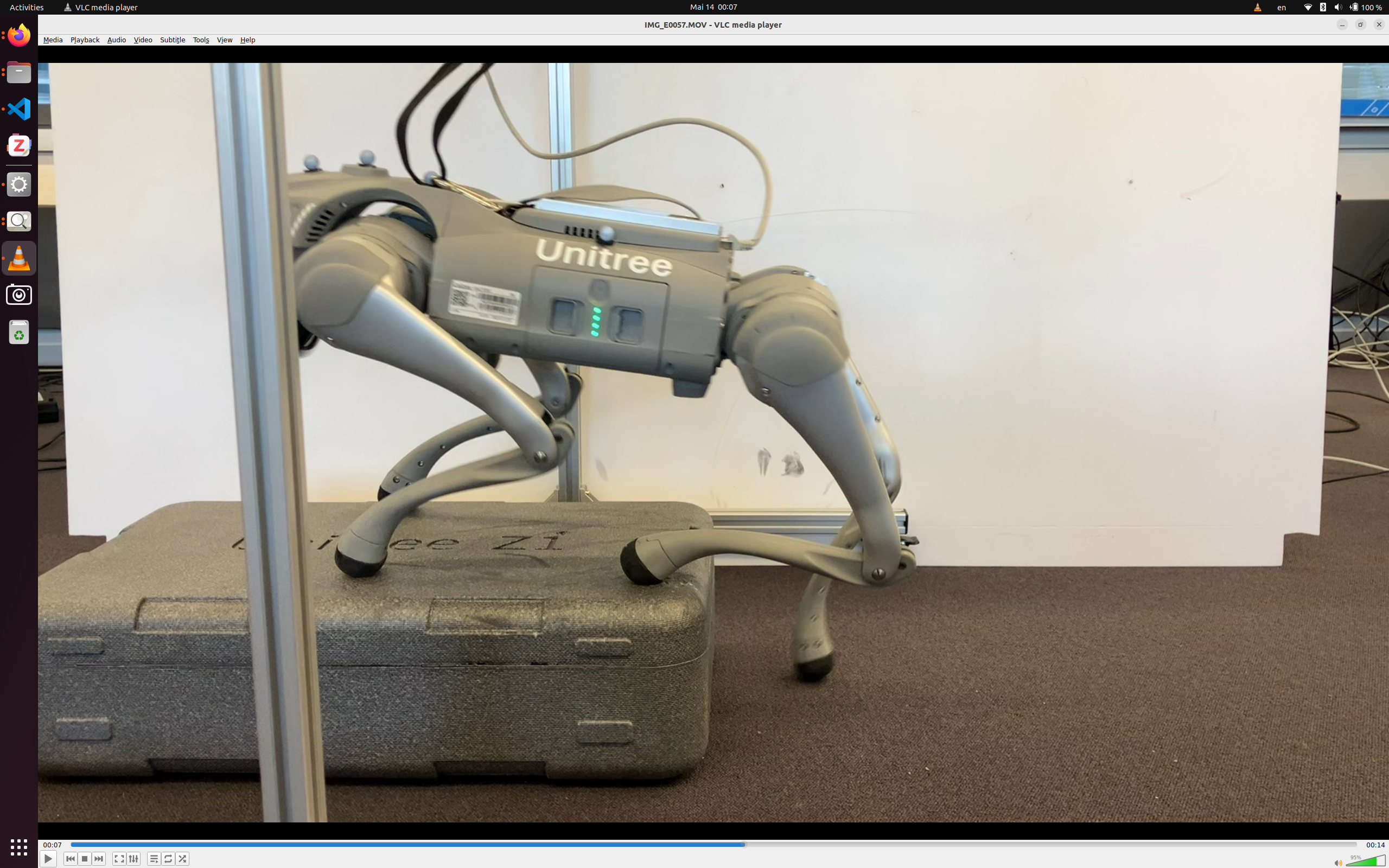}
        }
    }
    \hspace{-5mm}
    \subfloat{
        \reflectbox{
            \includegraphics[width=0.45\linewidth, trim=4cm 3cm 5cm 4.5cm, clip]{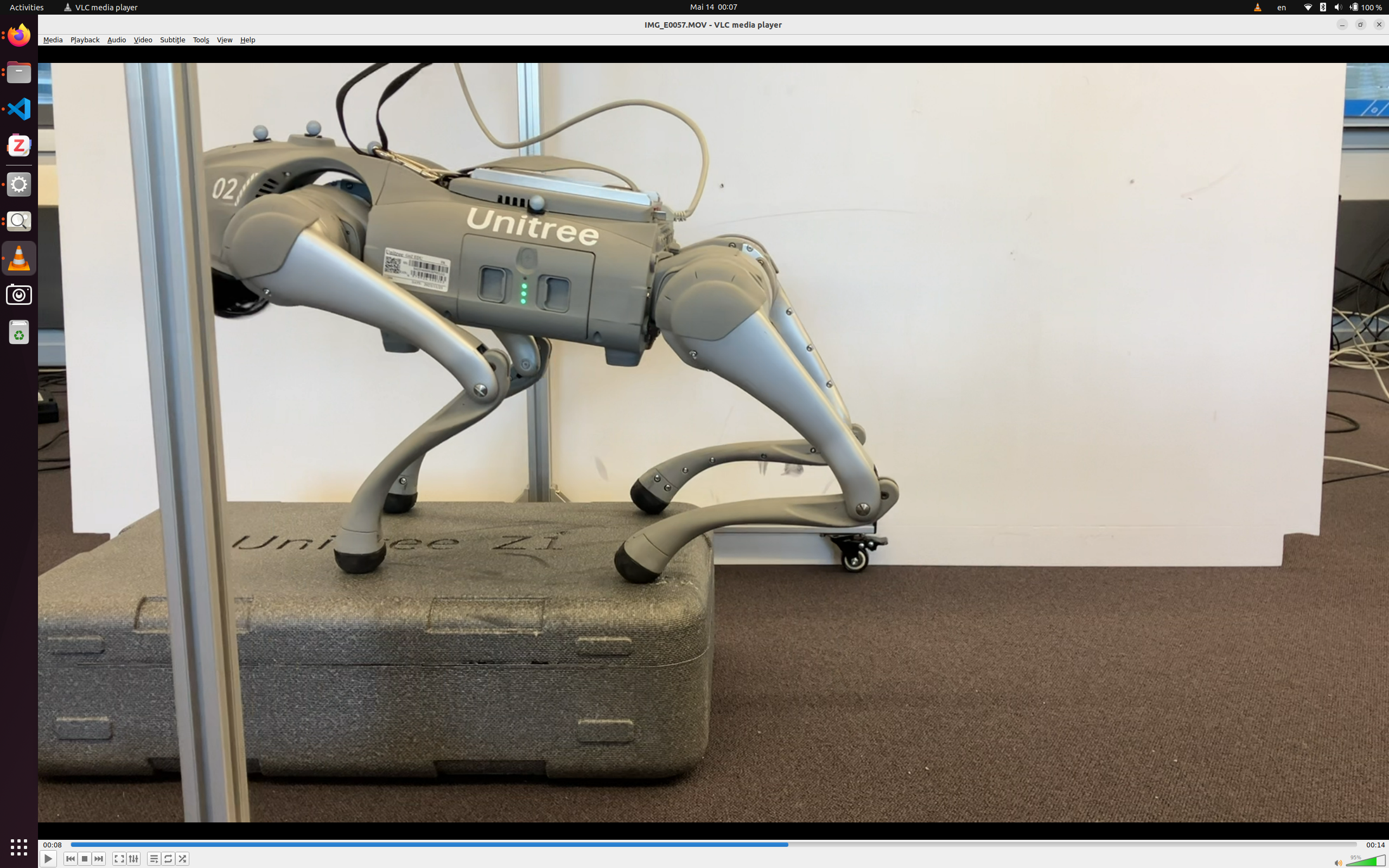}
        }
    }

    \caption{Successful transfer of a contact plan found by our pipeline to real hardware.}
    \label{fig:real_world_climb_box}
    \vspace{-4mm}
\end{figure}

Contact-explicit approaches, on the other hand, consider the contact planning explicitly. Early approaches cast the problem as a mixed-integer quadratic program (MIQP) \cite{deits2014footstep,aceituno2017simultaneous,ponton2021efficient,acosta2025perceptive}. However, the combinatorial complexity of the search over discrete contact variables makes the optimization problem computationally demanding. To circumvent this, recent works have solved a relaxed L1 minimization problem \cite{tonneau2020slim}, which has enabled online contact patch selection for quadruped locomotion \cite{corberes2023perceptive}. However, to cast the problem as MIQP, it is required to simplify the dynamics to be linear \cite{ponton2021efficient} (or consider no dynamics \cite{deits2014footstep}), which limits the range of possible behaviors that can be generated using MIQP formulation.

Monte-Carlo tree search (MCTS)  has recently emerged as an interesting alternative to MIQP. It has been shown in \cite{amatucci2022monte} that MCTS can reduce the computation time of MIQP significantly, while marginally sacrificing the optimality. Later, \cite{taouil2024non} combined MCTS with supervised learning to enable real-time contact sequence adaptation on flat terrains. However, both \cite{amatucci2022monte,taouil2024non} still used a simplified model of the dynamics and only optimized the contact sequence, and did not tackle the problem of automatic contact patch selection. Recently, \cite{dh2024diffusion,akizhanov2024learning} used MCTS together with a nonlinear MPC to enable agile locomotion on highly constrained surfaces such as stepping stones. However, they considered pre-defined gaits that can be limiting for a general case of locomotion in challenging environments.

\begin{figure*}[ht]
    \centering
    \includegraphics[width=0.99\textwidth]{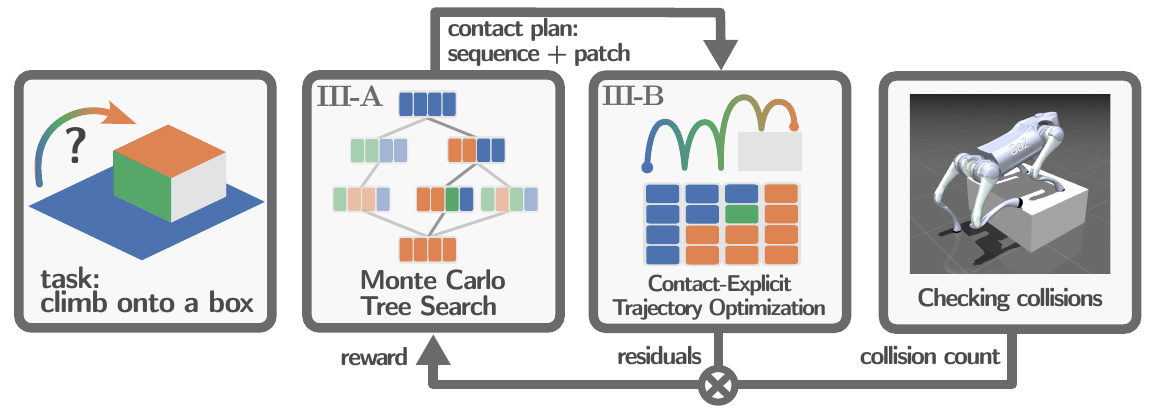}
    \vspace{-4mm}
    \caption{Overview of the proposed framework.}
    \vspace{-5mm}
    \label{fig:framework}
\end{figure*}

In this paper, we propose a  scalable formulation that can optimize for simultaneous contact sequence and patch selection while considering the full nonlinear dynamics of legged robots. The main contributions of the paper are summarized below:

\begin{itemize}
    \item We propose a formulation based on MCTS and whole-body trajectory optimization (TO) to simultaneously plan for dynamically feasible contact sequences and patches in highly constrained environments. 
    \item Through extensive simulation studies, we show that our framework can quickly find a variety of different solutions. This variety of solutions opens the future research to use advances in diffusion-based imitation learning to enable policies that can online select different contact plans in a feedback loop.
    \item We execute the generated plans in an MPC fashion on a real quadruped robot. To the best of our knowledge, this is the first demonstration of simultaneous contact sequence and patch selection using the whole-body dynamics of a quadruped. 
    \item We show that using the same framework, we can automatically generate highly complex acyclic behaviors on a humanoid robot.
\end{itemize}
The rest of the paper is structured as follows. In Section \ref{sec:preliminaries}, we present the fundamentals required to develop our framework. In Section \ref{sec:method}, we give a detailed description of our formulation. In Section \ref{sec:results}, we show the results and discuss them. Finally, in Section \ref{sec:conclusion}, we conclude our findings and outline the potential future research directions.

\section{Preliminaries}\label{sec:preliminaries}
\subsection{Contact-explicit trajectory optimization}\label{sec:preliminaries_ocp}
The dynamics of legged robots under rigid contact can be written in the following form
\begin{subequations}\label{eq:complementarityanddynamics}
    \begin{align}
        &\mathbf{M} \mathbf{\dot v} + \*b = \*S^\top \boldsymbol{\tau} +\sum_i \*J_{c,i}^\top \boldsymbol{\lambda}_i, \label{eq:dynamics}\\
        &0 \leq \Phi_i(\*q) \perp \boldsymbol{\lambda}_i \in \mathcal{K}_\mu \label{eq:complementarity}
    \end{align}
\end{subequations}

where $\*q \in \mathbb{SE}(3) \times \mathbb{R}^{n_q}$ and $\*v \in \mathbb{R}^{n_v}$ parameterize the configuration space and its corresponding tangent space. $\*M \in \mathbb{R}^{n_v \times n_v}$ is the mass matrix, $\*b \in \mathbb{R}^{n_v}$ is the vector grouping nonlinear terms including Coriolis, centrifugal, and gravitational terms, $\*S \in \mathbb{R}^{n_q \times n_v}$ is the selection matrix, $\boldsymbol \tau \in \mathbb{R}^{n_v}$ is the vector of actuating torques, $\*J_{c,i} \in \mathbb{R}^{3 \times n_v}$ is the Jacobian of the $i$th contact point, and $\boldsymbol \lambda_i \in \mathbb{R}^3$ is the contact force at the $i$th contact point. $\Phi_i(\*q)$ is the signed distance function of the $i$th end-effector relative to the environment. $\mathcal{K}_\mu=\{\boldsymbol{\lambda}\in\mathbb{R}^3|\lambda^z\geq\mu\norm{\boldsymbol{\lambda}^{x,y}}\}$ is the friction cone where $\mu$ is the friction coefficient. Equation \eqref{eq:complementarity} is the complementarity constraint that ensures the exertion of forces only at the points that are in contact with the environment. 

Equation \eqref{eq:complementarity} renders the dynamics of a legged robot hybrid and makes the trajectory optimization problem challenging to solve. To perform trajectory optimization, the contact-implicit formulation directly includes \eqref{eq:complementarityanddynamics} as constraints in the optimization problem \cite{posa2014direct}. Such MPCCs are highly challenging to solve and are prone to poor local minima \cite{nurkanovic2020limits}. The alternative formulation that we use in this work follows the contact-explicit formulation, where the problem is cast as a mixed-integer optimization problem \cite{aceituno2017simultaneous,ponton2021efficient}. In this formulation, activation of contact is explicitly encoded using integer variables (which end-effector makes contact with which surface), explicitly resolving the complementarity constraint \eqref{eq:complementarity}. In this work, we use a combination of Monte Carlo Tree Search (MCTS) and continuous trajectory optimization to solve the general contact-explicit trajectory optimization problem efficiently.

\subsection{Monte Carlo Tree Search}\label{sec:mcts_formulation}
    
MCTS formalizes a search problem by constructing a search graph $\mathcal{G} = (\mathcal{V}, \mathcal{E})$, where the set of nodes $\mathcal{V}$ contains the visited states and the set of edges $\mathcal{E}$ contains the visited transitions ${(\stateMCTS \overset{\actionMCTS}{\rightarrow}\stateMCTS ')}$.
Each transition maintains the state-action value $Q(\stateMCTS, \actionMCTS) \in \mathbb{R}$ and the number of visits $n(\stateMCTS, \actionMCTS) \in \mathbb{N}$. MCTS searches over this search graph iteratively using the following steps: the \textbf{Selection} phase begins at the root node (initial state) and successively chooses actions until a leaf node is reached (an unexpanded node or terminal state). If all children of a node are expanded, a child node is selected to be expanded based on a score that balances exploration and exploitation, e.g., upper confidence bound (UCB) \cite{kocsis2006bandit}. The \textbf{Expansion} phase adds the successor states to the graph by enumerating all possible actions, if the selected state is not terminal. The \textbf{Simulation} phase performs random actions from one of the successor states until a terminal node is reached and evaluates the reward over the full search path. Finally, the \textbf{Back-propagation} phase updates the state-action values and visit counts for all states along the selected and expanded nodes, using the reward obtained from the simulation.

\section{Method}\label{sec:method}
In this section, we outline our framework for simultaneously selecting a set of feasible contact sequences, patches and trajectories to reach a desired goal by a legged robot. An overview of the framework can be seen on Fig. \ref{fig:framework}.

\subsection{MCTS contact sequence and patch planning}\label{sec:method:mcts}

\subsubsection{Problem formulation}
We formulate contact sequence and patch selection as a Markov decision process (MDP).

In this MDP:
\begin{itemize}
    \item A state $\stateMCTS$ encodes, for each end-effector $i \in \llbracket 1, \Neeff \rrbracket$, its contact status $\contact_i \in \{0, 1\}$ and its associated contact surface (patch) $\patch_i \in \llbracket 1, \Npatch \rrbracket$ (only relevant when $i \in \mathcal{C}$, which denotes the set of end-effectors in contact). As defined, a state $\stateMCTS$ corresponds to a contact mode of the system. 
    \item An action $\actionMCTS$ defines the transition from one state to another, i.e., how each end-effector changes its contact: keeping, breaking, or making contact with a (possibly new) patch.
\end{itemize}

The trajectory is discretized into $N$ nodes, and for each node, we must determine the contact status and contact patch of each end-effector.
We denote the contact mode at step $k \in \llbracket 1, N \rrbracket$ as:
$\mode_k = ((\contact_{k, i})_{i \in \llbracket 1, \Neeff \rrbracket}, (\patch_{k, i})_{i \in \mathcal{C}_k})$.
Given an initial contact mode $\mode^i$ and a goal mode $\mode^g$, we aim to find a feasible sequence of contact modes $\mode_0^i \rightarrow \dots \rightarrow \mode^g_N$ within a limited number of steps. Once this full contact plan is found, all integer variables in the contact-explicit trajectory optimization problem are fixed and a gradient-based solver can then optimize for the continuous variables to ensure the existence of a whole-body trajectory consistent with all feasibility constraints (see Section \ref{sec:method:to}).

\subsubsection{Graph Pruning}

To avoid undesirable behaviors, such as frequent contact switches, we enforce a minimum contact duration: contact modes must remain constant over $\Nphase$ consecutive steps. We define the $j$-th contact phase as the set of steps $\varphi_j = \{j\Nphase, \dots, (j+1)\Nphase - 1\}$ during which the contact modes $(\mode_k)_{k \in \varphi_j}$ remain unchanged. Note that contact modes can be the same in two consecutive steps, extending the duration of a mode.

We simplify the search by considering only transitions between phases, up to phase $j_{\text{end}} = \left\lfloor N / \Nphase \right\rfloor$. This reduces the number of transitions to explore, making long-horizon planning more tractable. The search graph $\mathcal{G}$ is constructed such that the final contact mode matches the goal mode $\mode^g$.

To further reduce the size of the search space, we prune the following transitions from the graph:
\begin{itemize}
    \item Transitions where an end-effector changes patch without breaking contact, which is physically infeasible.
    \item Transitions that could result in crossing leg configurations, such as when the left leg's patch is to the right of the right leg’s patch.
\end{itemize}

Finally, we impose a minimum number of end-effectors in contact, denoted by $\mincnt$. Depending on the task, this can be set to zero to allow jumping motions.

\subsubsection{MCTS simulation policy}\label{sec:method:mcts:rollout}

In standard MCTS used for game play, the \textbf{Simulation}  phase performs random rollouts until a terminal state is reached.
In our approach, we replace these random simulations with a greedy heuristic policy that selects actions reducing the distance to the goal patches.
This heuristic guides the search toward shorter, more direct paths to the goal, similar to an $A^*$ strategy. As a result, it avoids inefficient back-and-forth contact transitions and improves planning efficiency.

\subsubsection{Reward formulation}\label{sec:method:mcts:reward}

Once a full contact sequence is found in the simulation phase, we run a whole-body TO (detailed formulation in Section \ref{sec:method:to}) to evaluate the physical and geometrical consistency of the plan. As the optimization problem is highly non-convex, we cannot guarantee zero constraint violations. In particular, self-collision and collision with the environment introduce many poor local minima.

To address this, we handle collision avoidance at two levels. In the TO, we enforce collision constraints only at the end-effector level to reduce complexity. This is detailed in Sections \ref{sec:method:to:eeff_ori} and \ref{sec:method:to:swing_cost}. In MCTS, we give a higher reward $R$ to contact sequences that lead to trajectories with fewer collisions (such as self-collisions and collisions of other body parts to the environment).

The reward is defined as $R = R_\text{col}R_\text{res}$, with:
\begin{align}\label{eq:reward}
    R_\text{col} &= \exp(\frac{-\alpha_\text{col}}{N} \sum_{k=1}^N \text{collision\_count}(\pos_k))
    \\
    R_\text{res} &= \sigma(-\alpha_\text{res}\log(\Pi_\text{res}))
\end{align}
$\alpha_\text{col} \in \mathbb{R}_{+}$ and $\alpha_\text{res} \in \mathbb{R}_{+}$ are scaling hyperparameters. $\sigma$ denotes the sigmoid function. "$\text{collision\_count}$" counts the number of self-collisions and collisions with the environment for a given configuration of the trajectory. $\Pi_\text{res}$ denotes the product of the following residual terms:

\begin{itemize}
    \item $r_\text{stat} \in \mathbb{R}_+$: the infinity norm of the KKT stationarity residual.
    \item $r_\text{eq} \in \mathbb{R}_+$: the residual of equality constraints (dynamics).
    \item $r_\text{ineq} \in \mathbb{R}_+$: the residual of inequality constraints.
    \item $r_\text{comp} \in \mathbb{R}_+$: the residual of complementarity conditions (note that this is not contact complementarity, but the complementarity constraints in the KKT condition due to the inequality constraints).
    \end{itemize}

\subsection{Trajectory Optimization}\label{sec:method:to}


For trajectory optimization, we use a formulation similar to the kino-dynamic whole-body trajectory optimization scheme in \cite{dai2014whole}. Let $\boldsymbol{\lambda}=[\boldsymbol{\lambda}_1;\cdots;\boldsymbol{\lambda}_{n_c}]\in\mathbb{R}^{3\Neeff}$ be the stacked forces at the contact points, $\*h \in \mathbb{R}^{9}$ be the vector of centroidal states. Defining $\*x=[\*h^\top~\*q^\top~\*v^\top]^\top$, $\*u=[\mathbf{\dot v}^\top~\boldsymbol{\lambda}^\top]^\top$, $\mathcal{C}=\{i\in\llbracket1,n_{ee}\rrbracket|\contact_i=1\}$, for a given initial state $\*x_0$, the optimal control problem (OCP) can be formulated as
\begin{subequations}
\begin{align}
    \min_{\tiny\substack{\{\*x_k, \*u_k\} \\ k\in\llbracket1, N\rrbracket}}& \quad  \sum_{k=1}^{N} L(\*x_k, \*u_k) \label{eq:ocp_cost_generic} \\
    \text{s.t.} 
    &\begin{bmatrix}
        \*h_{k}\\
        \*q_{k}\\
        \*v_{k}\\
    \end{bmatrix} = \begin{bmatrix}
        \*h_{k-1} + \*F_k\delta t\\
        \*q_{k-1} \oplus \*v_k \delta t\\
        \*v_{k-1} + \dot{\*v}_k \delta t\\
    \end{bmatrix},\label{eq:ocp_semi_implicit_euler}\\
    & \*h_k=\*A(\*q_k)\*v_k, \label{eq:ocp_cmm_constr} \\
    & \boldsymbol\tau_{\min} \preceq \boldsymbol{\tau}\preceq\boldsymbol\tau_{\max},\label{eq:ocp_torque_constr}\\
    & \forall i \in \mathcal{C},\nonumber\\
    & \quad\Phi_i(\*q_k)=0, {}_{\patch_i}\dot{\*r}_{c,i}^{x,y}=\*0, \label{eq:ocp_contact_constr_kin} \\
    & \quad\boldsymbol{\lambda}_i\in \mathcal{K}_\mu, \label{eq:ocp_contact_constr_force} \\
    & \forall i \notin \mathcal{C}, \nonumber\\
    & \quad \Phi_i(\*q_k)>0, \label{eq:ocp_contact_constr_swing} \\
    & \quad \boldsymbol{\lambda}_i=0, \label{eq:ocp_contact_constr_swing_force} \\
    & \text{Task Constraints}. \label{eq:ocp_task_constr}
\end{align}
\end{subequations}

where $\preceq$ denotes element-wise $\leq$ in \eqref{eq:ocp_torque_constr}. Equation \eqref{eq:ocp_cost_generic} is the generic cost, \eqref{eq:ocp_semi_implicit_euler} is the implicit Euler integration of the dynamics and whole-body kinematics ($\oplus$ denotes integration in the tangent space),  \eqref{eq:ocp_cmm_constr} is the centroidal momentum dynamics where $\*A$ is the centroidal momentum matrix. $\*F$ is the centroidal wrench and can be written down in terms of decision variables as follows

\begin{equation}
    \*F:=\begin{bmatrix} \sum_{i=1}^{\Neeff}\boldsymbol{\lambda}_i + m\*g \\ \sum_{i=1}^{\Neeff} (\*r_{c,i}-\*r_\text{CoM})\times \boldsymbol{\lambda}_i \end{bmatrix}
\end{equation}
where $\*r_{c,i},\*r_\text{CoM}$ are respectively the position of $i$th end-effector and the center of mass of the robot.
As joint torques are not decision variables, it is replaced in \eqref{eq:ocp_torque_constr} as a function of decision variables

\begin{equation}
    \boldsymbol{\tau}=\mathbf{M}^a \mathbf{\dot v} + \*b^a -\sum_i \*J_{c,i}^{a \top} \boldsymbol{\lambda}_i
\end{equation}
where superscript $a$ stands for the actuated part of the dynamics.
Different constraints are enforced depending on the contact status of end-effector $i$:
\begin{itemize}
    \item For points in contact: non-sliding contact kinematics constraint \eqref{eq:ocp_contact_constr_kin} and friction cones \eqref{eq:ocp_contact_constr_force};
    \item For points not in contact: foot-above-surface constraint \eqref{eq:ocp_contact_constr_swing} and and zero-force constraint \eqref{eq:ocp_contact_constr_swing_force}.
\end{itemize}
Task-specific requirements (\ref{eq:ocp_task_constr}), all enforced over the trajectory time domain and vary in different cases. In the following, we detail the costs and constraints that we enforce at the end-effector level to ensure collision avoidance with the environment. For simplicity, we will use $Q$ to denote cost weights.


\subsubsection{Contact phase}\label{sec:method:to:eeff_ori}
\begin{figure}
    \centering
    \includegraphics[width=0.8\linewidth]{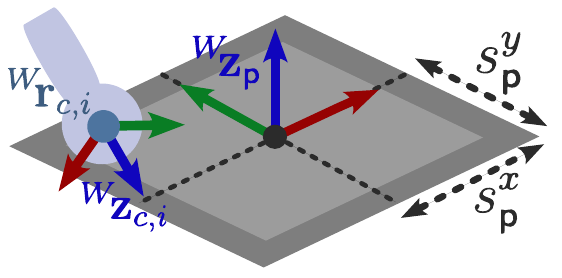}
    \vspace{-4mm}
    \caption{Geometrical representation of a rectangular contact patch of half-size $(s^x_{\patch}, s^y_{\patch})$. The light gray area is at a distance $\epsilon_{\patch}$ from the border. $_{W}\mathbf{z}_{c, i}(\pos) \in \mathbb{R}^3$ is the normal of the end-effector link frame.}
    \label{fig:patch_notation}
\end{figure}

We model each contact patch $\patch \in \llbracket 1, \Npatch \rrbracket$ as a rectangle with half-sizes $(s^x_{\patch}, s^y_{\patch})$ and plane normal $_{W}\mathbf{z}_{\patch}$ (see Figure~\ref{fig:patch_notation}).
Let $_{\patch_i}\mathbf{r}_{c, i} \in \mathbb{R}^3$ be the position of the $i$th contact point expressed in the frame of its patch $\patch_i$. $_{\patch_i}\mathbf{r}^{x}_{c,i}$ denotes the position along the $x$ axis (similarly for $y$ and $z$).

To ensure the contact point stays inside the patch boundaries, we apply the constraint:
\begin{align}
    |_{\patch_i}\mathbf{r}_{c, i}^{x, y}| \preceq s^{x, y}_{\patch_i}
\end{align}

To avoid contacts too close to the patch edges, we add a penalty cost:
%
\begin{align}\label{eq:eeff_border_cost}
    L_\text{edge}^{x, y} = \norm[\bigg]{\sigma\bigg(\beta_\text{edge}\bigg(
    \frac{|_{\patch_i}\mathbf{r}_{c, i}^{x, y}|}
         {s^{x, y}_{\patch_i} - \epsilon_{\patch_i}}
         - 1
     \bigg)\bigg)}^2_{Q_\text{edge}}
\end{align}
where:
\begin{itemize}
    \item $\beta_\text{edge} \in \mathbb{R}_+$ controls the stiffness of the penalty
    \item $\epsilon_{\patch_i} \in \mathbb{R}_+$ is a safety margin from the patch edge
    \item $\sigma$ is the sigmoid function
\end{itemize}

This cost is high near the edges and small when the contact point is well inside the patch. It encourages contact exactly at the center, while still keeping it away from the edges.

Another undesirable behavior is contacts made from the wrong side of a patch, e.g., a foot making contact from below the surface.
The following constraints and costs attempt to prevent this behavior.

For the $i$th end-effector in contact with $\patch_i$, let $_{W}\mathbf{z}_{c, i} \in \mathbb{R}^3$ denote the normal vector of the end-effector link frame (i.e., z-axis basis of the rotation matrix), as shown in Fig. \ref{fig:patch_notation}.
The sign of $\xi_i := \langle -_{W}\mathbf{z}_{c, i} , _{W}\mathbf{z}_{\patch_i} \rangle$ indicates if the end-effector points towards the contact surface from above ($\xi > 0$), which is desirable, or from below ($\xi_i < 0$).
We enforce the following condition to ensure contacts happen from the correct top side of the patch:
\begin{align}
    \quad \xi_i \geq 0.
\end{align}
Additionally, we encourage the end-effector to align with the contact surface normal using the cost:
\begin{align}
    L_{\xi} = \norm{1-\xi_i}^2_{Q_\alpha}
\end{align}

\subsubsection{Swing phase}\label{sec:method:to:swing_cost}
Hard constraints for collision avoidance are often difficult to define in complex environments. They also tend to produce trajectories that lie exactly on the constraint boundaries, leaving no margin for modeling errors and ultimately making real-world deployment challenging.

To address this, we use costs to attempt avoiding collisions during the swing phase of end-effectors. Two such cost terms are defined:
\begin{subequations}
    \begin{align}
        L_\text{sw}^z &=
        \begin{cases}
        \norm[\bigg]{
            \dfrac{\beta_{+}}{\beta_{+} + \left(_{\patch_i}\mathbf{r}^z_{c,i}\right)^4} 
            }^2_{Q_\text{sw}^z}
            & \text{if } _{\patch_i}\mathbf{r}^z_{c,i} \geq 0 \\
        \norm[\big]{
            \left(\beta_{-} \hspace{0.75mm} _{\patch_i}\mathbf{r}^z_{c,i} \right)^2 + 1
        }^2_{Q_\text{sw}^z}
            & \text{if } _{\patch_i}\mathbf{r}^z_{c,i} < 0
        \end{cases} \label{eq:under_plane_cost} \\
        L^{x, y}_\text{sw} &= \norm[\big]{\hspace{0.75mm} |_{\patch_i}\dot{\mathbf{r}}^{{x, y}}_{c,i}| \hspace{0.5mm} |_{\patch_i}\dot{\mathbf{r}}^{z}_{c,i}| \hspace{0.75mm}}^2_{Q_\text{sw}^{x, y}} \label{eq:swing_vel_cost}
    \end{align}
\end{subequations}

\begin{figure}
    \centering
    \includegraphics[width=0.9\linewidth]{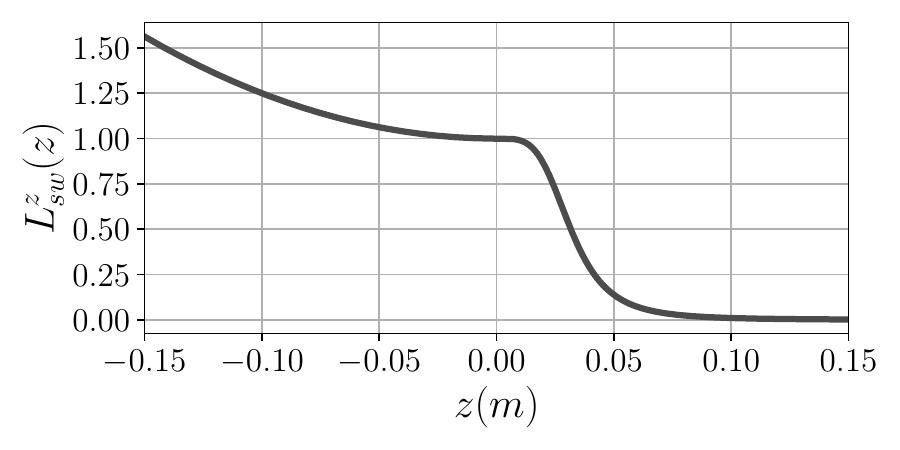}
    \vspace{-5mm}
    \caption{Cost $L_\text{sw}^z$ with $\beta_{-} = 5$ and $\beta_{+} = 10^{-7}$, as set in the experiments.}
    \label{fig:under_cost_graph}
\end{figure}

The first cost \eqref{eq:under_plane_cost} penalizes the end-effector being under the next contact plane. It remains strictly positive even when the height is zero, encouraging the swing motion to distance the surface. For positive heights, the cost quickly decreases, preventing overly high swings. The parameter $\beta_{+}$ sets the swing height, while $\beta_{-}$ penalizes being below the next contact patch. A graph of the cost function can be seen in Fig.~\ref{fig:under_cost_graph}.

The second cost \eqref{eq:swing_vel_cost} penalizes large horizontal velocities $_{\patch_i}\dot{\mathbf{r}}^{x, y}_{c,i}$ when the vertical velocity $_{\patch_i}\dot{\mathbf{r}}^{z}_{c,i}$ is also large. This is particularly helpful when stepping up onto higher surfaces, such as boxes or stairs. When combined with \eqref{eq:under_plane_cost}, the end-effector is encouraged to first go up above the next contact plane, then move laterally toward the patch, reducing the chance of hitting the patch edge.

The reference patch $\patch_i$, in which the end-effector position $_{\patch_i}\mathbf{r}_{c,i}$ is expressed, changes halfway through the swing phase. In the first half, the end-effector position is expressed relative to the previous contact patch (the one it just break contact with), and in the second half, relative to the upcoming contact patch.
This switching of reference frames enables smooth swing trajectories especially when the patches have different orientations.

\section{Results}\label{sec:results}

In this section, we present the results of applying the proposed framework to a Go2 quadruped robot for two dynamic locomotion tasks: crossing a gap (see Fig. \ref{fig:task_cross_gap}) and climbing onto a box (see Fig. \ref{fig:task_climb_box}). In section \ref{sec:results_sim}, we show that our framework can find a diverse set of whole-body collision-free trajectories within seconds. In section \ref{sec:results_real_world}, we show that contact plans found by our framework successfully transfer to the real world. Finally, in section \ref{sec:humanoid_result}, we show that the proposed contact planning algorithm can also be applied to a humanoid robot.


We implemented the trajectory optimization framework using the sequential quadratic programming (SQP) of Acados \cite{Verschueren2021} with Pinocchio \cite{carpentier2019pinocchio} and CasADI \cite{Andersson2019} bindings. We used MuJoCo \cite{todorov2012mujoco} to visualize the trajectories and count the collisions in the tree search. We also use MuJoCo as a physics engine to test later the execution of the plans in an MPC fashion, before testing it on the real robot.

The solver optimizes a state trajectory and action plan over $N = 50$ steps. The time horizon is set to $3s$, which corresponds to a time step of $\Delta t = 0.06s$.
The solver performs $75$ SQP iterations in total, which is enough to reach convergence in most cases. With those settings, on a 13th Gen Intel® Core™ i9-13900H, the solver takes around $1.1s$ to optimize the full trajectory.

\subsection{Tasks description}\label{sec:tasks_description}

\subsubsection{Crossing a gap}
The first task, illustrated in Fig. \ref{fig:task_cross_gap}, consists of crossing a gap of parameterized length $l_\text{gap}$ and width $w_\text{gap}$. We added two lateral walls of parametrized inclination $\alpha_{gap}$. This task has $\Npatch=4$ patches : the plane on which the robot starts, the two walls and the plane across the gap. The goal is to have the four end-effectors on the plane across the gap.

\subsubsection{Climbing onto a box}
\begin{figure}[h]
    \centering
    \subfloat[\scriptsize Crossing a gap ($l_\text{gap} = 0.9$)]{
        \includegraphics[width=0.49\linewidth]{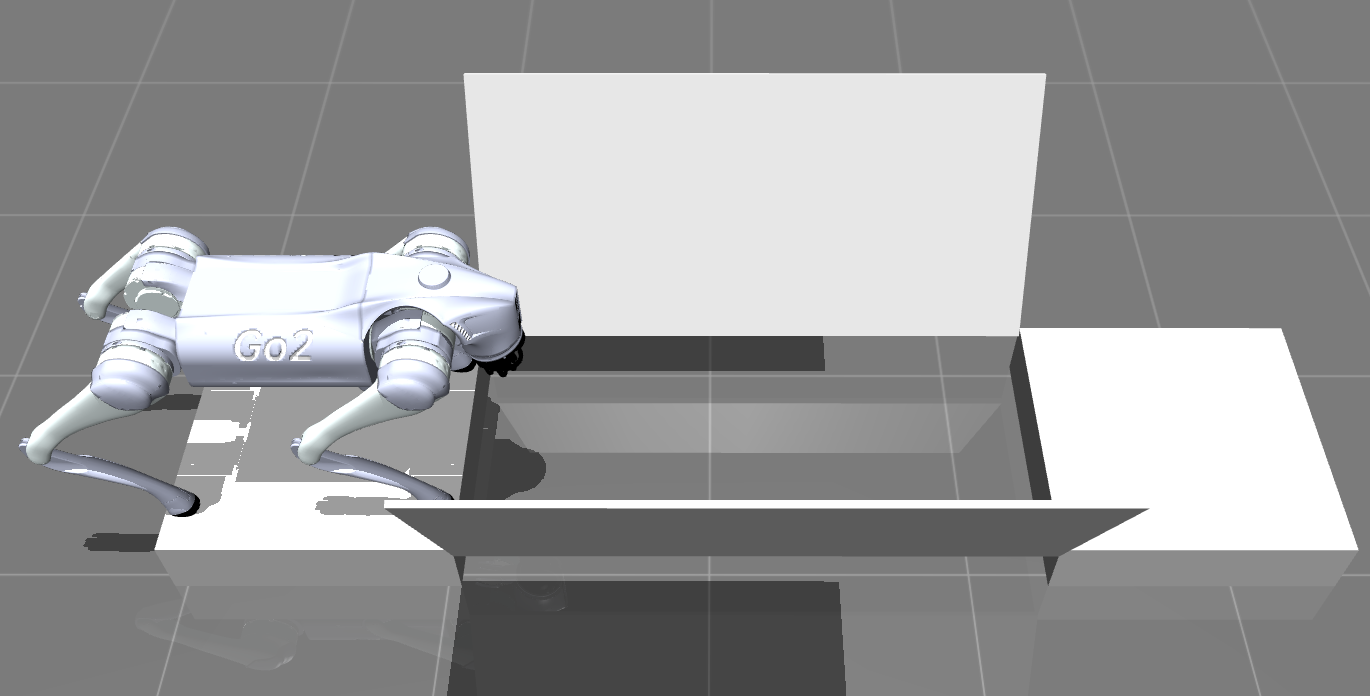}
        \label{fig:task_cross_gap}
    }
    \subfloat[\scriptsize Climbing onto a box ($h_\text{box} = 0.4$)]{
        \includegraphics[width=0.49\linewidth]{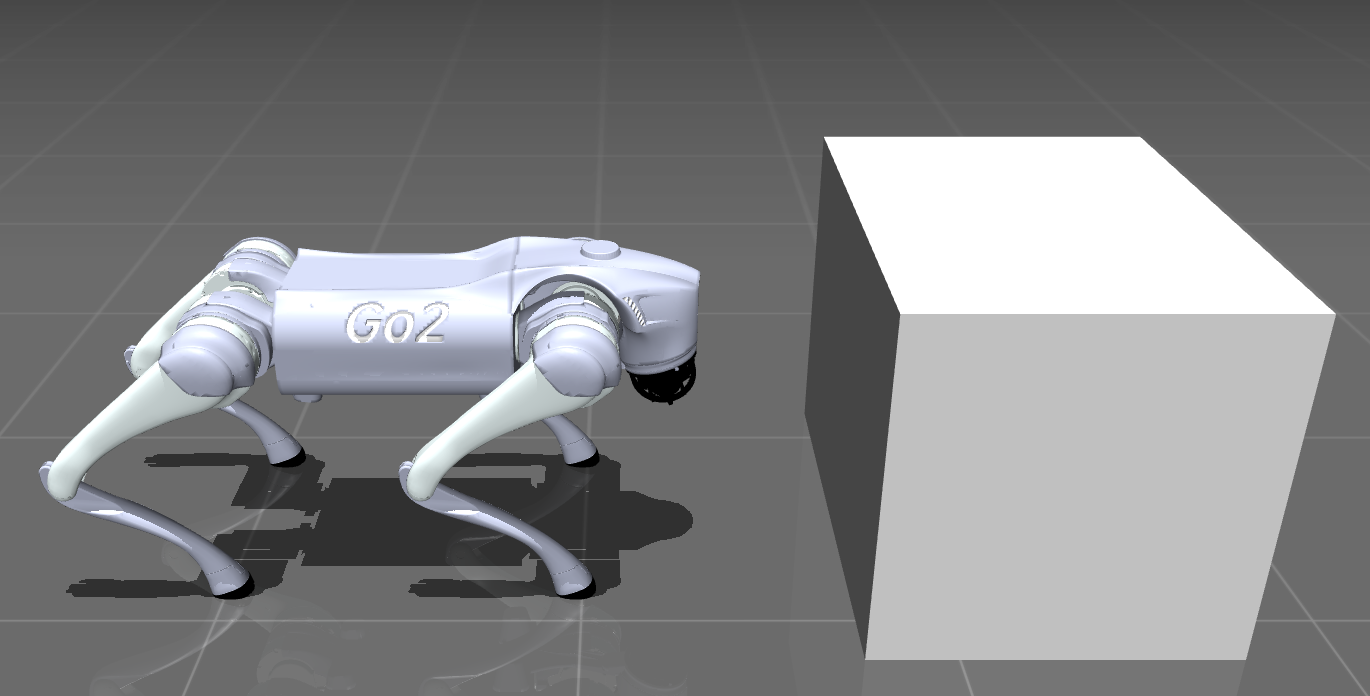}
        \label{fig:task_climb_box}
    }
    \caption{The two scenarios considered in this work, displayed in MuJoCo simulator.}
    \label{fig:task_environments}
\end{figure}

 The second task, illustrated in Fig. \ref{fig:task_climb_box}, consists of climbing onto a box of parameterized dimensions $(l_\text{box}, w_\text{box}, h_\text{box})$. The center of the box is offset in the $x$ direction by $x_\text{off} = 0.5\text{m}$ to leave some space in between the box and the robot.
 We only considered $\Npatch=3$ different patches: the floor, the patch of the box facing the robot, and the top face of the box. The goal is to have the four end-effectors on the top face of the box.

We considered different parametrizations of the tasks with increasing difficulty to test our framework on, those can be seen on Table \ref{tab:task_parameters}.
 
\begin{table}[h]
\centering
\caption{Task parameters for contact planning}
\begin{tabular}{|l|c|c|}
\hline
\textbf{Parameters} & \textbf{Crossing a gap} & \textbf{Climbing onto a box} \\
\hline
$\Npatch$ & 4 (start, $2$ walls, across) & 3 (floor, front, top) \\
$n_\varphi$ & 5 & 4 \\
$\mincnt$ & 0 (allows free flying) & 2 (no free flying) \\
fixed param. & \scriptsize $w_{gap} = 0.4, \alpha_{gap}= 65^{\circ}$ & $l_{box} = 0.4, w_{box} = 0.5$ \\
variable param. & $l_\text{gap}$ & $h_\text{box}$ \\
values & $[0.3, 0.5, 0.7, \dots , 1.5]$ & $[0.2, 0.3, 0.4, \dots, 0.7]$ \\

\hline
\end{tabular}
\label{tab:task_parameters}
\end{table}

\subsection{Statistical analysis}\label{sec:results_sim}


For both tasks, we let MCTS run for $1000$ iterations.
We used an exploration-exploitation parameter $C=1$ in the UCB. The scaling parameters of the reward in (\ref{eq:reward}) are set to $\alpha_\text{col}=0.1$ and $\alpha_\text{res}=1/3$. Statistics are computed over $5$ different runs of the same experiment.

\begin{figure}[h]
    \centering
    \subfloat[Crossing a gap]{
        \includegraphics[width=0.48\linewidth]{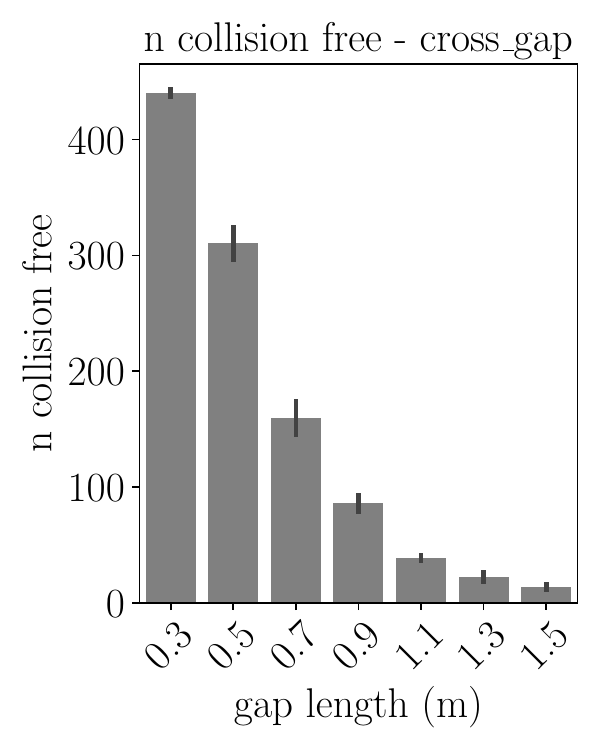}
        \label{fig:cross_gap_collision_free}
    }%
    \subfloat[Climbing onto a box]{
        \includegraphics[width=0.48\linewidth]{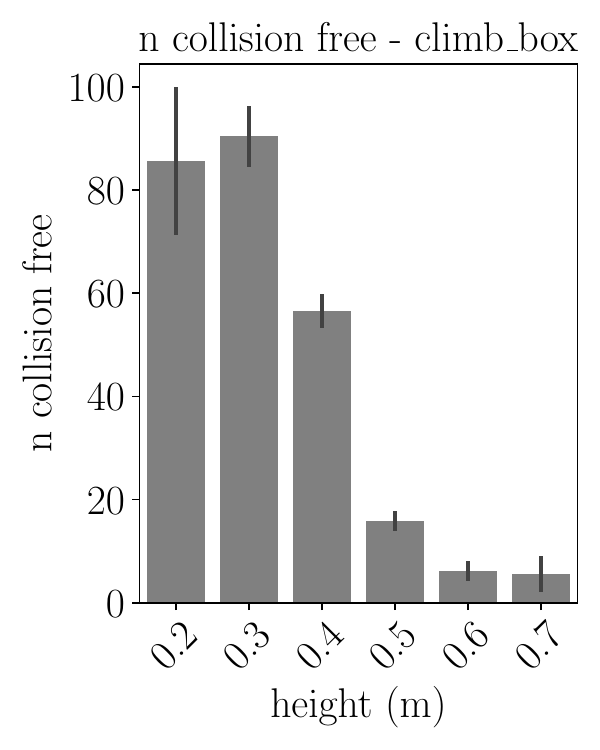}
        \label{fig:climb_box_collision_free}
    }
    \caption{Number of collision-free trajectories across different task parameters. The thin line represents the standard deviation across the different runs.}
    \label{fig:collision_free_comparison}

\end{figure}

\begin{figure*}
    \centering
    \subfloat{
        \includegraphics[width=0.98\textwidth]{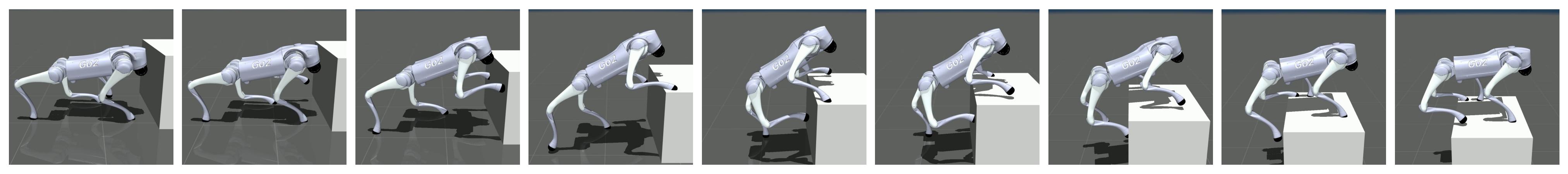}
    }
    \\
    \vspace{-4mm}
    \subfloat{
        \includegraphics[width=0.98\textwidth]{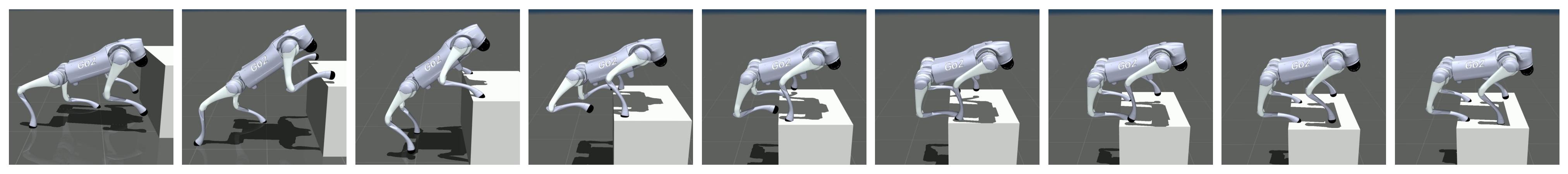}
    }
    \\
    \vspace{-4mm}
    \subfloat{
        \includegraphics[width=0.98\textwidth]{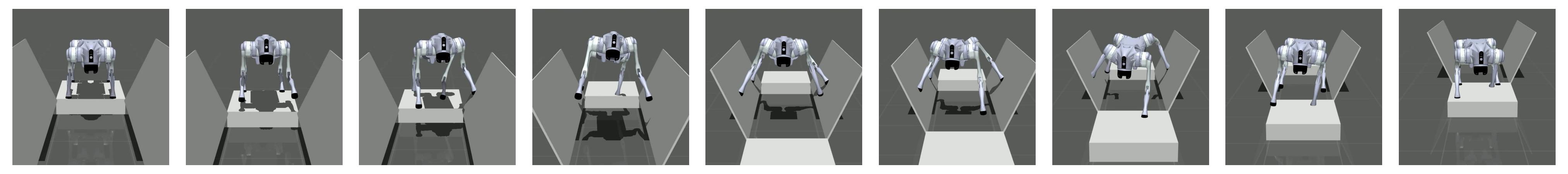}
    }
    \\
    \vspace{-4mm}
    \subfloat{
        \includegraphics[width=0.98\textwidth]{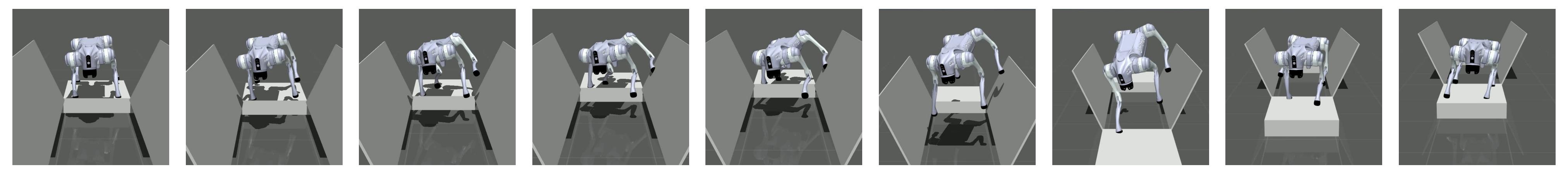}
    }
    \\
    \vspace{-3mm}
    \caption{Quaduped robot climbing onto a $0.4$m high box (top) and crossing a $1.1$m long gap (bottom). Snapshots of some collision-free trajectories.}
    \label{fig:go2_traj}
\end{figure*}

The proposed framework successfully generates a diverse set of whole-body, collision-free trajectories for both tasks. Figure~\ref{fig:collision_free_comparison} reports the number of such trajectories found as a function of box height and gap length. As expected, the number of feasible solutions decreases with task difficulty. Nonetheless, the framework consistently finds valid trajectories for box heights up to 70cm and gap lengths up to 1.5m.
Multiple examples of successful trajectories can be seen in Fig. \ref{fig:go2_traj} and in the accompanying video.

Fewer successful trajectories are observed for the box-climbing task, as it involves more challenging collision avoidance. Additionally, a minimum of two end-effectors in contact ($\mincnt = 2$) was enforced for this task, further limiting the solution space.

\begin{figure}[h]
    \centering
    \subfloat[Crossing a gap]{
        \includegraphics[width=0.49\linewidth]{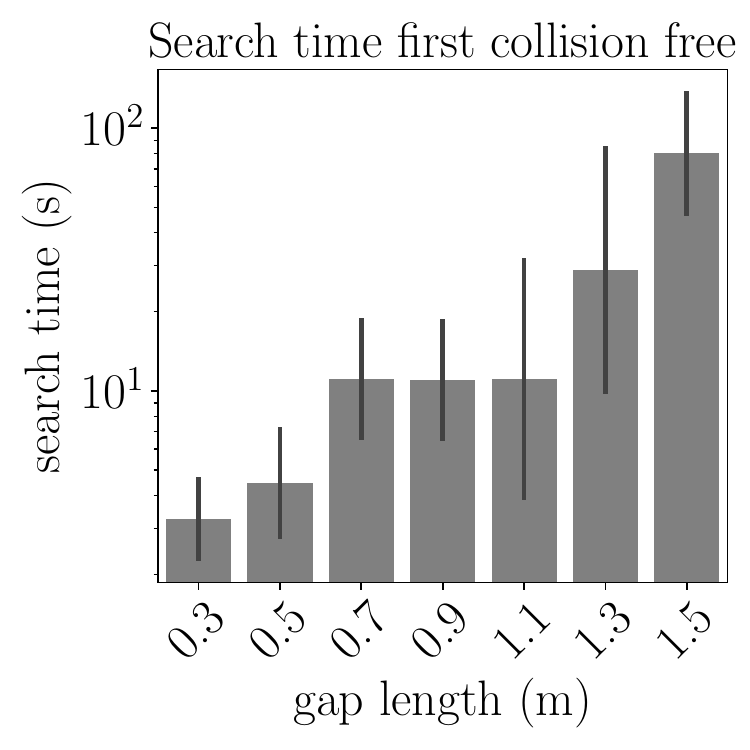}
        \label{fig:cross_gap_first_collision_free}
    }
    \subfloat[Climbing onto a box]{
        \includegraphics[width=0.49\linewidth]{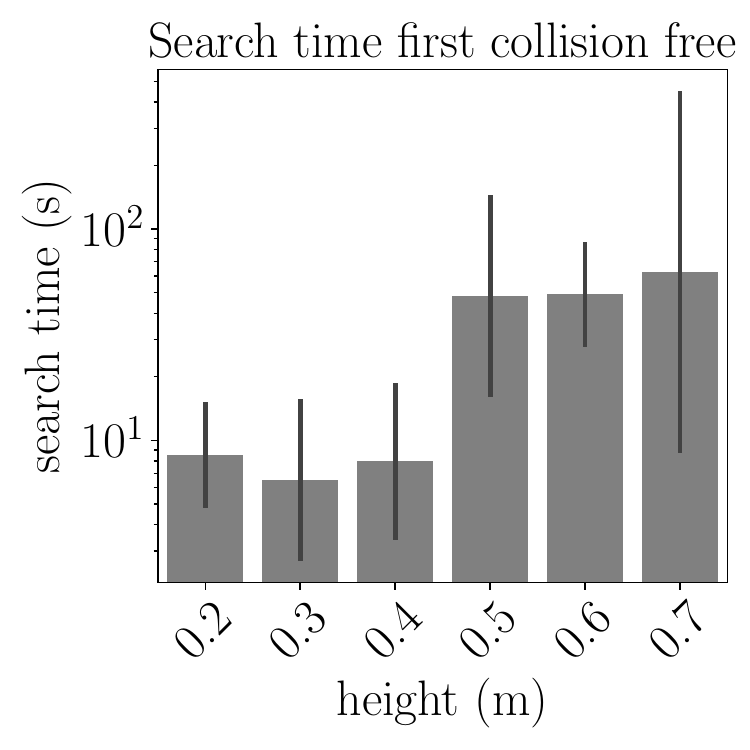}
        \label{fig:climb_box_first_collision_free}
    }
    
    \caption{Solving time for the first collision-free trajectory across different parameters. The thin line represents standard deviation across the different runs.}
    \label{fig:first_collision_free_solve_time}
\end{figure}

\begin{figure}[h]
    \centering
    \subfloat{
        \includegraphics[width=\linewidth]{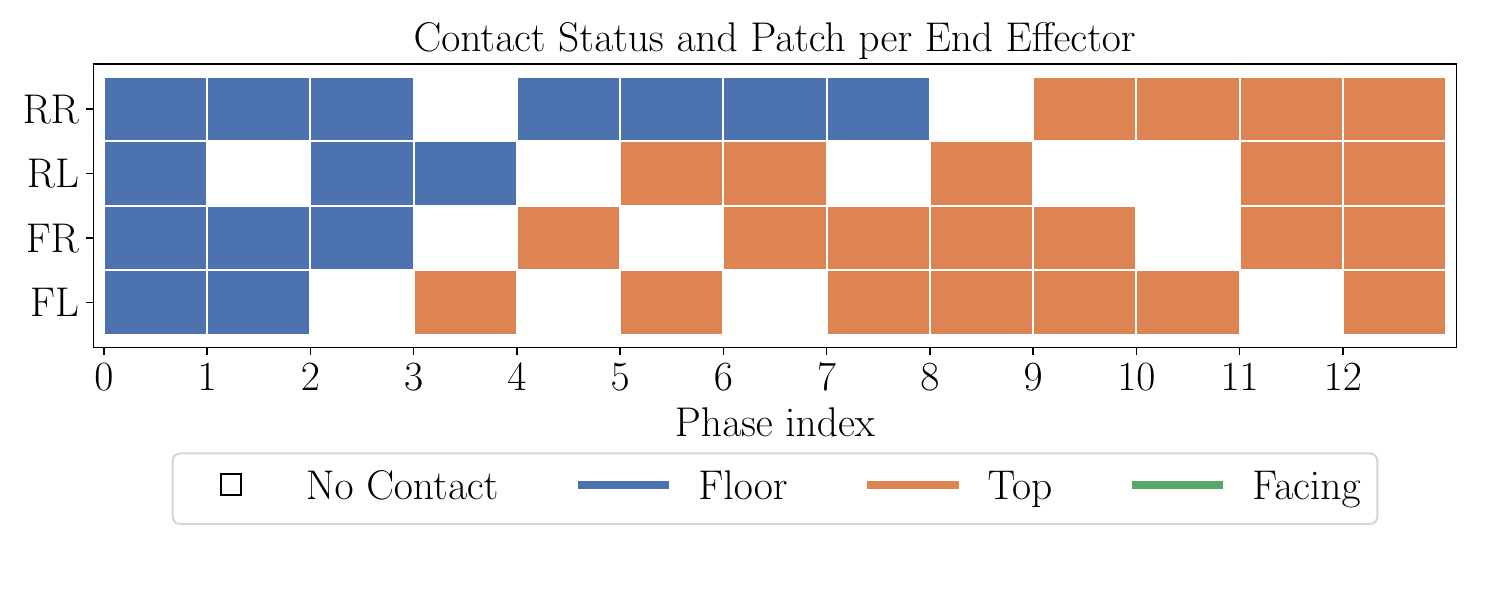}
        \label{fig:climb_box_contact_seq_1}
    }
    \vfill
    \vspace{-1.5cm}
    \subfloat{
        \includegraphics[width=\linewidth, trim=0 0 0 1cm,clip]{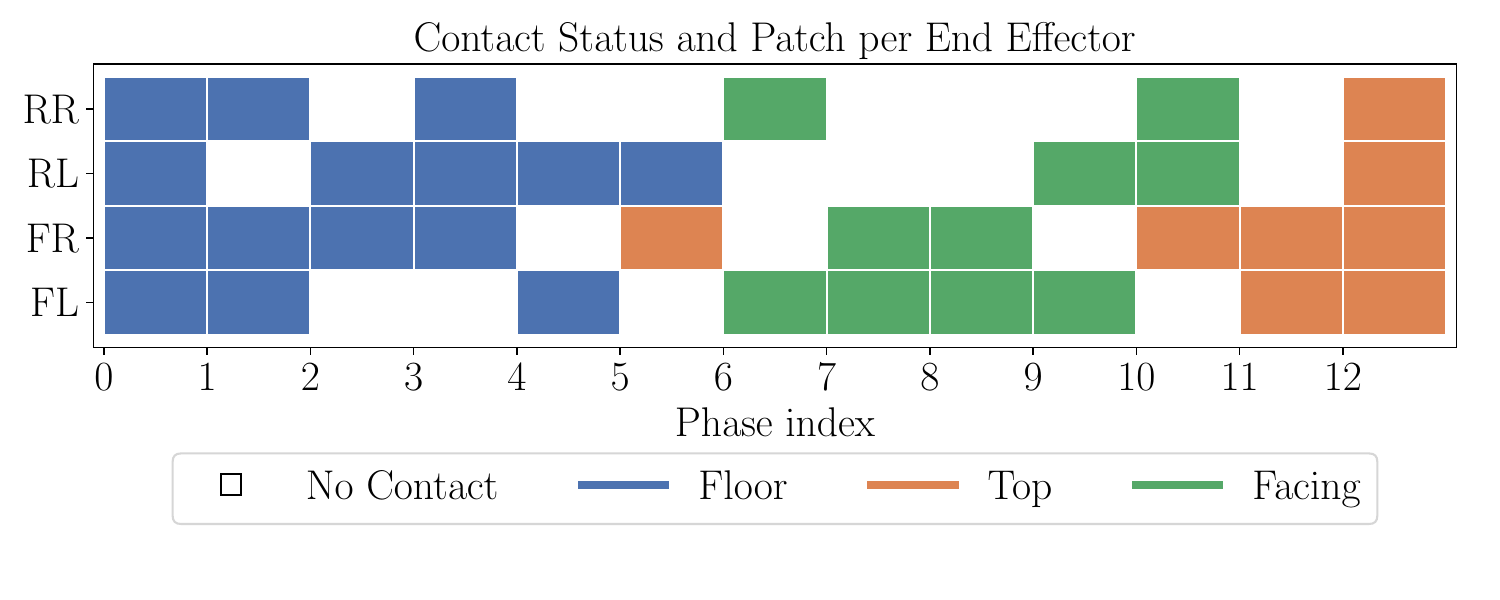}
        \label{fig:climb_box_contact_seq_2}
    }
    \vspace{-0.7cm}
    \caption{Two distinct contact plans for the climbing-box task with $h_{box} = 0.4\text{m}$. The contact patch facing the robot (in green) is used only in the contact plan below.}
    \label{fig:climb_box_two_contact_plans}
\end{figure}

The first collision-free trajectory is typically found in under 10 seconds for simpler scenarios ($h_\text{box} \leq 40$cm and $l_\text{gap} \leq 1.1$m). Extensive results can be found in Fig. \ref{fig:first_collision_free_solve_time}. 
The pipeline finds approximately 3 collision-free trajectories per minute for the box-climbing task and about 8 per minute for the gap-crossing task. These rates could be further improved through parallelization of the search. Finally, Fig. \ref{fig:climb_box_two_contact_plans} shows two different contact sequences of successful trajectories for a given task, showcasing the diversity of contact plans found.

\subsection{Real hardware experiments}\label{sec:results_real_world}

To show that the contact plans found by our framework can be successfully executed, we developed a closed-loop MPC and used it as a feedback controller both in the simulation and on the real hardware. The MPC relies on the same trajectory optimization formulation described in Section~\ref{sec:method:to}, with only minor adaptations: mainly a re-tuning of cost weights and a smaller discretization time step of $\Delta t = 0.03\text{s}$ and a short horizon. On our system, the MPC replans at approximately $30$Hz using SQP-based real-time iteration with asynchronous solver calls, meeting real-time performance requirements.

Figure~\ref{fig:real_world_climb_box} shows snapshots of a Unitree Go2 quadruped climbing a $0.18$m high box by executing one of the contact plans found in simulation. The full experiment can be seen in the accompanying video.

\subsection{Humanoid experiments}\label{sec:humanoid_result}

\begin{figure*}
    \centering
    \subfloat{
        \includegraphics[width=0.98\textwidth]{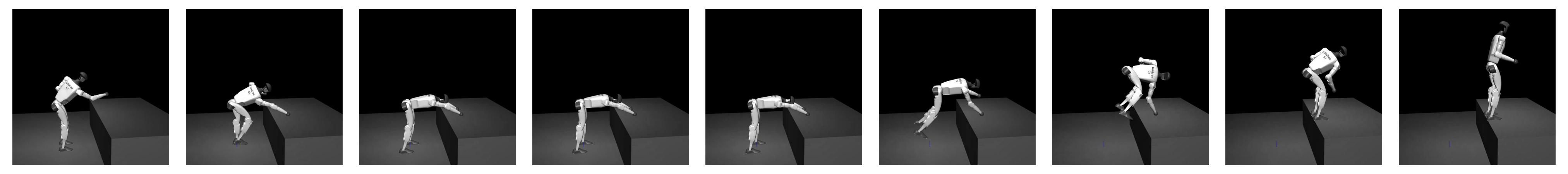}
    }
    \\
    \vspace{-4mm}
    \subfloat{
        \includegraphics[width=0.98\textwidth]{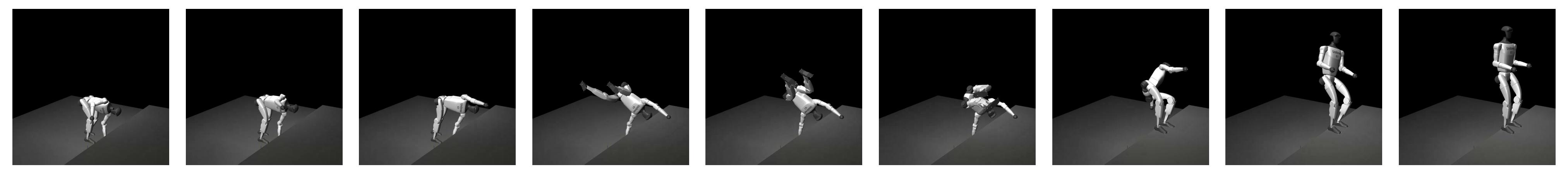}
    }
    \vspace{-3mm}
    \caption{Humanoid robot climbing onto a $0.5$m high box. Snapshots of two collision-free trajectories.}
    \label{fig:humanoid_climb_box}
\end{figure*}

The proposed contact planning framework is general enough to be directly applied to humanoid robots with minimal adjustments.  
By modifying the trajectory optimization (TO) formulation in Section~\ref{sec:method:to} to account for different contact interactions (surface-surface contact for the feet), our MCTS framework successfully finds $65$ collision-free trajectories of a humanoid robot climbing a $0.5$m high box within $300$ MCTS iterations. Dynamic feasibility of the trajectories is satisfied for most cases, but not all (the constraint residuals in some of the motions are not negligible).
The first collision-free solution is found after $52$ seconds.  
Figure~\ref{fig:humanoid_climb_box} shows two successful trajectories, highlighting the diversity of the solutions found.

\section{Conclusion and future work}\label{sec:conclusion}

In this paper, we proposed a complete pipeline for simultaneous contact and motion planning for multi-contact legged locomotion. Our proposed framework efficiently generates diverse, whole-body collision-free trajectories across a range of dynamic locomotion tasks and different robot morphologies.
We executed some of the contact plans on a quadruped robot using MPC with real-time iteration in both simulation and hardware. We also showed that our framework can generate agile complex multi-contact behaviors for a humanoid robot within tens of seconds of computation.

While the full pipeline is not fast enough for online contact replanning, it could be used to gather a large dataset of trajectories and contact plans for training generalist policies through imitation learning. We plan to address this in our future work.

\bibliographystyle{IEEEtran}
\bibliography{root}
\end{document}